*Engineering Note*

# mGPT: A Probabilistic Planner Based on Heuristic Search

**Blai Bonet**                                                    BONET@LDC.USB.VE
*Departamento de Computación*
*Universidad Simón Bolívar, Venezuela*

**Héctor Geffner**                                        HECTOR.GEFFNER@UPF.EDU
*ICREA & Universitat Pompeu Fabra*
*Paseo de Circunvalación 8, Barcelona 08003, Spain*

## Abstract

We describe the version of the GPT planner used in the probabilistic track of the 4th International Planning Competition (IPC-4). This version, called mGPT, solves Markov Decision Processes specified in the PPDDL language by extracting and using different classes of lower bounds along with various heuristic-search algorithms. The lower bounds are extracted from deterministic relaxations where the alternative probabilistic effects of an action are mapped into different, independent, deterministic actions. The heuristic-search algorithms use these lower bounds for focusing the updates and delivering a consistent value function over all states reachable from the initial state and the greedy policy.

## 1. Introduction

mGPT is a planner based on heuristic search for solving Markov Decision Processes (MDPs) specified in the high-level planning language PPDDL. mGPT captures a fragment of the functionality of the GPT system that handles non-determinism and incomplete information, in both qualitative and probabilistic forms, including POMDPs and Conformant planning (Bonet & Geffner, 2000).

mGPT supports several algorithms and admissible heuristic functions (lower bounds) that when combined generate a wide range of solvers. The main algorithms are `lrtdp` and `hdp`. Both are heuristic-search algorithms for solving MDPs that make use of lower bounds for computing a *consistent* value function $V$: a function with Bellman residuals bounded by a user-provided parameter $\epsilon$ over all states reachable from a given initial state $s_0$ and the greedy policy based on $V$ (Bonet & Geffner, 2003b, 2003a).

The lower bounds are derived by solving relaxations of the input problem. Since the algorithms for solving the relaxations are also based on heuristic search, we have implemented "stackable" software components that are created in sequence for computing complex heuristic functions from simpler ones.

## 2. Algorithms

We divide the algorithms into two groups: those that deliver consistent value functions with respect to a user-provided parameter $\epsilon$, and those that select actions in real time. The first class of algorithms compute an $\epsilon$-consistent value function $V$ for all states reachable from the initial state $s_0$, and the greedy policy $\pi_V$ based on $V$.

In the following subsection, we give definitions of admissible and consistent value functions, and greedy, partial and proper policies. Then, we present the algorithms implemented by mGPT.

## 2.1 Consistent Value Functions, and Greedy, Partial and Proper Policies

A value function $V$ is admissible if it is non-overestimating; i.e. if the value $V(s)$ at each state $s$ is a lower bound on the optimal expected cost of starting at $s$. $V$ is $\epsilon$-consistent at state $s$ if its Bellman residual at $s$,

$$R(s) \stackrel{\text{def}}{=} \left| V(s) - \min_{a \in A(s)} \left[ c(s,a) + \sum_{s' \in S} Pr(s'|s,a) V(s') \right] \right|, \tag{1}$$

is less than or equal to $\epsilon$. Here, $A(s)$ denotes the actions that are applicable at $s$, $c(s,a)$ is the cost of applying action $a$ in $s$, and $Pr(\cdot)$ is the probabilistic transition function. If $V$ is 0-consistent at $s$, then we say that $V$ is consistent at $s$.

A state $s$ is reachable from the initial state $s_0$ and policy $\pi$ if there exists a trajectory $s_0, s_1, \ldots, s_n$ such that $s_n = s$ and $P(s_{k+1}|s_k, \pi(s_k)) > 0$ for all $0 \le k < n$. In other words, if the state $s$ can be reached with positive probability from $s_0$ in zero or more steps using the policy $\pi$.

It is known that the greedy policy $\pi_V$ based on the value function $V$, defined as

$$\pi_V(s) \stackrel{\text{def}}{=} \operatorname*{argmin}_{a \in A(s)} \left[ c(s,a) + \sum_{s' \in S} Pr(s'|s,a) V(s') \right], \tag{2}$$

is optimal when $V$ is $\epsilon$-consistent over *all* states for a sufficiently small $\epsilon$. Yet, since our goal is to find an optimal policy with respect to the initial state $s_0$ and the states reachable from it, it is sufficient for $V$ to be admissible and $\epsilon$-consistent over the states that are reachable from $s_0$ and $\pi_V$.

A partial policy $\pi$ is a policy that doesn't need to be defined for all states. It is closed with respect to a state $s$ if $\pi$ is defined over $s$ and all states reachable from $s$ and $\pi$, it is proper with respect to $s$ if a goal state can be reached from every state reachable from $s$ and $\pi$, and finally it is proper if it is proper with respect to all states.

## 2.2 Algorithms that Compute $\epsilon$-Consistent Value Functions

For the first group of algorithms, mGPT implements Value Iteration (`vi`), Labeled Real-Time Dynamic Programming (`lrtdp`), and Heuristic Dynamic Programming (`hdp`).

Value Iteration (Bertsekas, 1995) is applied over the states that can be reached from the given initial state and the available operators, and yields an $\epsilon$-consistent value function over all of them.[1] mGPT's `vi` serves as a bottom-line reference for comparison with the other algorithms.

---

1. On undiscounted problems like those in probabilistic planning, some conditions are neeeded in order for VI to finish with an $\epsilon$-consistent value function (Bertsekas, 1995).

Labeled Real-Time Dynamic Programming (Bonet & Geffner, 2003b) is a heuristic-search algorithm that implements a labeling scheme on top of the `rtdp` algorithm (Barto, Bradtke, & Singh, 1995) to improve its convergence. `Lrtdp` works by performing simulated trials that start at the initial state and end at "solved" states, selecting actions according to the greedy policy $\pi_V$ and successor states according to their corresponding transition probabilities. Initially, $V$ is the input heuristic function, and the only solved states are the goal states. Then, each time an action is picked at state $s$, the value of $s$ is updated by making it consistent with the value of its successors. At the end of each trial, a labeling procedure is called that checks whether new states can be labeled as solved: a state is solved if its value and the value of all its descendents are $\epsilon$-consistent. The algorithm ends when the initial state is labeled solved. At that point all states reachable from the initial state $s_0$ and the greedy policy $\pi_V$ are $\epsilon$-consistent. The labeling mechanism also guarantees that $\pi_V$ is a proper partial policy with respect to $s_0$.

Heuristic Dynamic Programming (Bonet & Geffner, 2003a) is the second heuristic-search algorithm supported in mGPT for solving MDPs. `Hdp` performs systematic depth-first searches over the set of states reachable from the initial state $s_0$ and the greedy policy $\pi_V$ looking for $\epsilon$-inconsistent states and updating their values. On top of this search, a labeling scheme based on Tarjan's strongly-connected components procedure (Tarjan, 1972), identifies the states that are solved and that do not need to be revisited. The initial value function is given by a heuristic function, and the algorithm ends when the initial state is solved. As with `lrtdp`, the labeling mechanism guarantees that $\pi_V$ is proper with respect to $s_0$.

### 2.3 Algorithms for Real-Time Action Selection

The second class of algorithms do not attempt to solve the given MDP; they rather select actions in real-time after a limited amount of processing without offering any guarantees on the quality of the resulting policies. Algorithms in this group include an extension of the Action Selection for Planning algorithm (`asp`) (Bonet, Loerincs, & Geffner, 1997) for probabilistic domains, which is basically an `rtdp` algorithm with lookahead. `Asp`, like `rtdp`, performs value function updates over states and so cannot get trapped into a loop. Thus, although the policy delivered by `asp` is suboptimal, it is a proper policy; i.e. a policy that is *guaranteed* to reach a goal state.

## 3. Heuristics

All these algorithms assume that the initial value function is given by a heuristic function that provides good cost estimates, and in particular, `lrtdp` and `hdp` expect this heuristic to be admissible. As described by Pearl (1983), informative admissible heuristics can be obtained by solving suitable relaxations of the input problem. Two such relaxations are supported in mGPT: the min-min relaxation, and the Strips relaxation. The first defines a (deterministic) shortest-path problem in the original state space; the second is used to define (deterministic) shortest-path problems in atom space.[2] Thus, while the first is solved in time polynomial in the number of states, the shortest-path problems defined by the second are solved in time polynomial in the number of atoms. Both methods yield lower bounds on the expected cost to the goal from a given state, yet the bounds produced by the min-min relaxation are stronger than those produced by the Strips relaxation.

---

2. Atoms refer to the propositional symbols used in the representation language, PPDDL in our case, to define the problem. The number of atoms is polynomial in the size of the input, while the size of the state space is, in general, exponential in the number of atoms.

## 3.1 Min-Min State Relaxation

The idea behind the min-min relaxation is to transform the input probabilistic problem, described by its Bellman equations

$$V^*(s) \stackrel{\text{def}}{=} \min_{a \in A(s)} \left[ c(s,a) + \sum_{s' \in S} Pr(s'|s,a)V^*(s') \right], \tag{3}$$

into a deterministic shortest-path problem with Bellman equations of the form,

$$V^*_{min}(s) \stackrel{\text{def}}{=} \min_{a \in A(s)} c(s,a) + \min \left\{ V^*_{min}(s') : P(s'|s,a) > 0 \right\}. \tag{4}$$

At the level of the representation language, the min-min relaxation is built by transforming each probabilistic operator of the form:

$$o = \langle \varphi, [\, p_1 : \alpha_1, \ldots, p_n : \alpha_n \,] \rangle, \tag{5}$$

where $\varphi$ is the precondition of $o$ and each $\alpha_i$ is the $i$th probabilistic effect (with probability $p_i$), into a set of *independent and deterministic* operators of the form:

$$o_i = \langle \varphi, \alpha_i \rangle, \quad 1 \le i \le n. \tag{6}$$

Thus, in the min-min relaxation one can actually choose the most convenient non-deterministic effect of an operator, and hence, the cost of the relaxation is a lower bound on the expected cost of the original probabilistic problem.

The min-min relaxation is a deterministic problem that can be solved by means of standard path-finding algorithms. For example, it can be solved with Dijkstra's algorithm, A*, IDA*, or a deterministic version of `lrtdp` (i.e. a labeled `lrta` algorithm (Korf, 1990)).

mGPT provides two methods for computing the min-min heuristic from this relaxation: `min-min-ida*`, which uses IDA*, and `min-min-lrtdp`, which uses `lrtdp`. Both versions are *lazy* in the sense that the heuristic values of states are computed as needed when the planner requires them.

## 3.2 Strips Relaxation

The Strips relaxation in turn converts the deterministic problem obtained from the min-min relaxation into a Strips problem, and then obtains lower bounds for the original MDP by computing lower bounds for the resulting Strips problem using the methods developed in classical planning (e.g., Bonet & Geffner, 2001; Haslum & Geffner, 2000; Hoffmann & Nebel, 2001; Edelkamp, 2001; Nguyen & Kambhampati, 2000). These methods run in polynomial time in the number of atoms yet, unlike the min-min relaxation, require casting the min-min relaxation into Strips format, a conversion that, like the conversion of ADL into Strips (Gazen & Knoblock, 1997), may require exponential time and space (see below).

In mGPT, the Strips relaxation is obtained directly from the original problem, by first transforming the probabilistic operator into the form:

$$o = \langle \mathtt{prec}, [\, p_1 : (\mathtt{add}_1, \mathtt{del}_1), \ldots, p_n : (\mathtt{add}_n, \mathtt{del}_n)\,]\,\rangle, \tag{7}$$

where $\mathtt{prec}, \mathtt{add}_i, \mathtt{del}_i$ are *conjunctions* of literals that represents the precondition, the $i$th add list, and the $i$th delete list of operator $o$ respectively, and $p_i$ are probabilities that sum to 1. In order to take the operators into form (7), all disjunctive preconditions, conditional effects, and quantifiers are removed as described by Gazen and Knoblock (1997).

Once all operators have the form (7), the Strips relaxation is generated by splitting the operators into $n$ independent Strips operators of the form:

$$o_i = \langle \mathtt{prec}, \mathtt{add}_i, \mathtt{del}_i\rangle, \quad 1 \leq i \leq n. \tag{8}$$

The following heuristics are implemented in mGPT upon the Strips relaxation. The first two are lower bounds on the optimal cost of the Strips relaxation and hence on the optimal (expected) cost of the original MDP, the third one is not necessarily a lower bound on either cost.

- The $h^m$ heuristics (`h-m`) (Haslum & Geffner, 2000) are heuristics that recursively approximate the cost of achieving a set of atoms $C$ from the initial state by the cost of achieving the most costly subset of size $m$ in $C$. They are computed by a shortest-path algorithm over a graph with nodes standing for sets of at most $m$ atoms, and result in values $h^m(s)$ that estimate the cost of reaching a goal state from $s$. We use the option `h-m-k` in mGPT to refer to the $h^m$ heuristic with $m = k$.

- Pattern database heuristics (`patterndb`) (Edelkamp, 2001) compute optimal costs of relaxations of the Strips problem defined by some of the multi-valued variables that are *implicit* in the problem (e.g. the location of a block in the blocksworld domain is an implicit multi-valued variable whose possible values are either the table or the top of any other block). This heuristic is also precomputed only once, at the beginning, and provides a lower bound on the cost of an arbitrary state to the goal. A pattern database is computed by projecting the Strips problem with respect to a set of atoms $A$ (those that define the multi-valued variables) and then solving the resulting problem optimally with Dijkstra's algorithm. Multiple pattern databases can be combined either by taking max or sum. In the latter case, the pattern database is referred to as additive.[3] We use additive pattern databases as defined by Haslum, Bonet, and Geffner (2005) where some constraints of the original problem are preserved into the projection; something that often results in stronger heuristics. `Patterndb-k` refers to a pattern database heuristic defined by $k$ multi-valued variables.

- The FF (`ff`) heuristic implements the heuristic function used in the FF planner (Hoffmann & Nebel, 2001). It is computed by building the so-called *relaxed* planning graph and finding a plan in it. The heuristic is then the number of operators in such a plan.

---

3. Some conditions are required for adding two pattern databases such that the result remains admissible. A sufficient condition is that $A \cap B = \emptyset$ if the sets $A$ and $B$ are those used to build the projections respectively.

The relaxed planning graph is the version of the graph constructed by Graphplan (Blum & Furst, 1997) when delete lists are ignored. It can be shown that computing the `ff` heuristic can be done in polynomial time in the size of the input problem (Hoffmann & Nebel, 2001). This heuristic however is informative but non-admissible.

As it is shown below, these heuristics can be plugged directly into the planning algorithm or they can be used to compute more informative heuristics. For example, the `patterndb` heuristic can be used within IDA* to solve the min-min relaxation, which gives a stronger heuristic than the `patterndb` heuristic. Thus, mGPT implements algorithms and heuristics as stackable software components so that an element in the stack is used to solve the elements above it.

## 4. Implementation

This section gives some details on the implementation of mGPT together with examples on its use. The mGPT system is implemented in C++ upon a preliminary parser offered by the organizers of IPC-4.

### 4.1 Hash Tables

Perhaps the most important component of modern search-based planners is the internal representation of states and hash tables. Since mGPT uses different search algorithms and hash tables to solve a given instance (e.g. when more informative heuristics are computed from less informative ones), good internal representations and hash table implementation are critical for good performance.

After grounding all atoms and operators, a state is represented by the ordered list of the atoms that hold true in the state. A state $s$ can appear associated with different data in multiple hash tables simultaneously. Thus, instead of having multiples "copies" of $s$, mGPT implements a system-wide *state-hash-table* that stores the representation of the states referenced in all hash tables so that entries in such tables simply contain a reference into the state-hash-table. In this way, the planner saves time and space.

Another issue that has large impact on performance is the average number of collisions in each hash table. Two points are relevant for keeping the number of collisions low: the hashing function and the size of the hash table. For the former, we have seen that cryptographic hashing functions like MD4 behave very well even though they are slower than more traditional choices. For the latter, mGPT uses hash tables whose size is equal to a large prime number (Cormen, Leiserson, & Rivest, 1990).

### 4.2 Algorithms and Heuristics

Each algorithm in mGPT is implemented as a subclass of the abstract `algorithm` class whose members are a reference to a problem and, in some cases, a reference to a hash table and a parameter $\epsilon$. Similarly, each heuristic in mGPT is implemented as a subclass of the abstract `heuristic` class whose members are a reference to a problem and a function that maps states to non-negative values. Simple heuristics like the constant-zero function are straightforward, others like `min-min-lrtdp` are implemented by a class whose members are, in addition to above, references to a hash table and to an `lrtdp` algorithm.

### 4.3 Examples

The main parameters on a call to mGPT are "`-a <algorithm>`" that specifies the algorithm to use, "`-h <heuristic>`" that specifies the heuristic function, and "`-e <epsilon>`" that specifies the threshold $\epsilon$ for the consistency check. A typical call looks like:

```
mGPT -a lrtdp -h h-m-1 -e .001 <domain> <problem>
```

which instructs mGPT to use the `lrtdp` algorithm with the `h-m-1` heuristic and $\epsilon = 0.001$ over the domain and problem files specified.

The `h-m-1` heuristic is admissible but very weak. The following example shows how to compute the `min-min-lrtdp` heuristic using `h-m-1` as the base heuristic:

```
mGPT -a lrtdp -h "h-m-1|min-min-lrtdp" -e .001 <domain> <problem>
```

The pipe symbol is used to instruct the planner how heuristics are to be computed using other heuristics.

Another possibility is to use mGPT as a reactive planner in which decisions are taken on-line with respect to a heuristic function that is improved over time. For example,

```
mGPT -a asp -h ff <domain> <problem>
```

uses the `asp` algorithm with the `ff` heuristic, while

```
mGPT -a asp -h "zero|min-min-ida*" <domain> <problem>
```

uses the `asp` algorithm with the `min-min-ida*` heuristic computed from the constant-zero heuristic. Other combinations of algorithms and heuristics are possible. mGPT also accepts parameters to control initial hash size, a weight on the heuristic function, values for dead-end states, verbosity level, lookahead settings for `asp`, etc.

## 5. The Competition

The competition suite consisted of 7 probabilistic domains named blocksworld, exploding-blocksworld, boxworld, fileworld, tireworld, towers-of-hanoise, and zeno. Blocksworld and exploding-blocksworld are variations of the standard blocksworld domain for classical planning. Boxworld is a logistics-like transportation domain. Fileworld is a file/folder domain where the uncertainty is only present at the initial situation where the destination of each file is set. Tireworld and towers-of-hanoise are variations of the classical tireworld domain and towers-of-hanoi. Zeno is a traveling domain with a fuel resource.

Some of the domains come in two variations: a goal-oriented version where the goal is to be achieved with certainty while minimizing expected costs, and a reward-oriented version that involves rewards. The mGPT planner handles the first type of tasks only.

In the competition we used the `lrtdp` algorithm with the `patterndb-1` heuristic, a parameter $\epsilon = 0.001$, and a weight $W = 5$ for the heuristic function. In some cases, when the `patterndb-1` heuristic was too poor, the planner switched automatically to the `asp` algorithm with the `ff` heuristic.

| problem name | runs | failed | successful | time | reward |
|---|---|---|---|---|---|
| `blocksworld-5` | 30 | 0 | 30 | 43 | 494.1 |
| `blocksworld-8` | 30 | 0 | 30 | 60 | 487.7 |
| `blocksworld-11` | 30 | 0 | 30 | 130 | 465.7 |
| `blocksworld-15` | 30 | 0 | 30 | 7,706 | 397.2 |
| `blocksworld-18` | — | — | — | — | — |
| `blocksworld-21` | — | — | — | — | — |
| `exploding-bw` | — | — | — | — | — |
| `boxworld-c5-b10` | 30 | 0 | 30 | 6,370 | 183.6 |
| `boxworld-c10-b10` | — | — | — | — | — |
| `boxworld-c15-b10` | — | — | — | — | — |
| `fileworld-30-5` | 30 | 0 | 30 | 2,220 | 57.6 |
| `towers-of-hanoise` | — | — | — | — | — |
| `tireworld-g` | 30 | 14 | 16 | 48 | 266.6 |
| `tireworld-r` | 30 | 0 | 30 | 39 | 0 |
| `zeno` | 30 | 0 | 30 | 162 | 500 |

Table 1: Results for the mGPT planner over the competition problems. The table shows problem name, number of runs, number of failed and successful runs (see text), and time and reward averages. A dash means that mGPT was not able to solve the problem. Times are in milliseconds.

## 5.1 Results

The competition was held through a client/server model. Each planner was evaluated in each problem over a number of runs under supervision of the server. The planner initiated the session by connecting to the server and then interacted with it by exchanging messages. Each run consisted of actions sent by the planner whose effects were transmitted back from the server to the planner. Thus, the current state of the problem was maintained both by the planner and the server.

Table 1 shows the results for mGPT over the competition problems. For each problem, 30 runs were executed. The table shows the number of runs, the number of failed runs (i.e. those that finished without reaching a goal state), the number of successful runs (i.e. those that finished at goal states), and the time and reward averages per run.[4] For the blocksworld, the problem `blocksworld-xx` means a problem with `xx` blocks, for boxworld, the problem `boxworld-cxx-byy` means a problem with `xx` cities and `yy` boxes.

As it can be seen from the table, mGPT did not solve `exploding-bw`, the larger instances of `blocksworld` and `boxworld`, and it also failed on approximately half of the instances in `tireworld-g`. The difficulties encountered by mGPT in solving these problems often had not so much to do with the probabilities involved, but with the domains, and in particular, with the encodings. The basic algorithms used by mGPT try to solve the problems by computing a value function with $\epsilon$-residuals over the relevant states (those reachable from the initial state by an optimal policy). For this, mGPT computes an admissible heuristic function by solving either the min-min relaxation, the Strips relaxation, or both. A problem faced by this approach is that in many instances neither of these relaxations could be solved. Here, we give a detailed explanation of the problems encountered by mGPT over the different domains. It is worth noting that many of these difficulties would surface in any Strips planner as well, even if the probabilities are ignored.

---

4. The competition format was reward-based while our presentation here is cost-based. It is straightforward to go from one format to the other.

- Blocksworld and exploding blocksworld: the operator encodings have preconditions containing universally-quantified negative literals, as the result of not using a 'clear' predicate. For example,

```
(:action pick-up-block-from
  :parameters (?top - block ?bottom)
  :precondition (and (not (= ?top ?bottom))
                     (forall (?b - block) (not (holding ?b)))
                     (on-top-of ?top ?bottom)
                     (forall (?b - block) (not (on-top-of ?b ?top))))
  :effect (and (decrease (reward) 1)
               (probabilistic
                 0.75 (and (holding ?top) (not (on-top-of ?top ?bottom)))
                 0.25 (when (not (= ?bottom table))
                        (and (not (on-top-of ?top ?bottom))
                             (on-top-of ?top table)))))
)
```

This complex encoding is not standard in planning and makes our atom-based heuristics almost useless. mGPT could solve the instances with 5, 8, 11 and 15 blocks but not those with 18 and 21 blocks. For exploding blocksworld, mGPT was unable to solve it as the parser is incomplete and does not parse some complex constructs.

- Boxworld: the encoding contains a 'drive-truck' operator that moves the truck to its intended destination with probability 0.8 and to one of three "wrong destinations" with probability 0.2/3 each. The encoding specifies the unintended effects by means of nested conditional effects of the form

```
(:action drive-truck
  :parameters (?t - truck ?src - city ?dst - city)
  :precondition (and (truck-at-city ?t ?src) (can-drive ?src ?dst))
  :effect (and (not (truck-at-city ?t ?src))
               (probabilistic
                 0.2 (forall (?c1 - city)
                       (when (wrong-drive1 ?src ?c1)
                         (forall (?c2 - city)
                           (when (wrong-drive2 ?src ?c2)
                             (forall (?c3 - city)
                               (when (wrong-drive3 ?src ?c3)
                                 (probabilistic
                                   1/3 (truck-at-city ?t ?c1)
                                   1/3 (truck-at-city ?t ?c2)
                                   1/3 (truck-at-city ?t ?c3)))))))))
                 0.8 (truck-at-city ?t ?dst)))
)
```

Our Strips relaxation, like any planner that converts ADL-style operators into Strips, suffers an exponential blow up in this domain: with 10 cities, there are more than a thousand operators for each grounded ADL-operator. This set included problems with 5, 10 and 15 cities.

- Fileworld: in this domain, there are 30 files that need to be filed into one of 5 different folders: the exact destination determined probabilistically. The optimal policy for this problem, and any proper policy, must *prescribe an action for more than* $5^{30}$ *states*, all of them relevant. The consequence is a problem with millions of relevant states that need to be stored into the hash table if the task is to compute a proper policy. The `patterndb-1` heuristic for this problem is not informative, as revealed by an analysis of the values stored in the pattern database, and thus mGPT switched automatically to the `asp` algorithm with the `ff` heuristic.

- Towers-of-hanoise: as in the blocksworld domain, the encoding is complex with operators that have disjunctions and universally-quantified negative literals in the preconditions, and complex conditional effects. Yet the problem that prevented mGPT from solving any problem in this domain is a bug in the code that implements conditional effects which did not surface in the other domains.

- Tireworld: there are two versions: a goal-based version called `tireworld-g` and a reward-based version called `tireworld-r`. The domain contains multiple dead ends at locations where the car gets a flat tire and no spare tire is available. Some of the dead ends are unavoidable; i.e. there is no proper policy for this problem. All trials for the reward-based version end successfully since there is no requirement to reach a goal position, rather the objective is to maximize the accumulated reward. mGPT treated both versions as goal-based problems as it does not deal directly with reward-based problems.

## 6. Conclusions

The mGPT planner entered into the probabilistic planning competition combines heuristic-search algorithms with methods for obtaining lower bounds from deterministic relaxations. The results obtained at the competition were mixed with some of the difficulties having to do with the selection of domains and encodings which do not match the capabilities of mGPT: mGPT tries to compute proper solutions using heuristics derived from the Strips relaxations. As we have described, some of the domains could not be solved due to the number of relevant states, and others due to the complexity of the Strips relaxations themselves.

For the definition of good benchmarks for MDP solvers, it is crucial to define what constitutes a solution and what is the bottom line for assessing performance. In classical planning, for example, the solutions are plans and the bottom line is given by blind-search algorithms; progress in the field can then be measured by the distance to this bottom line. In the probabilistic setting, this is more difficult as it is not always clear what it means to solve a problem. This, however, needs to be defined in some way, otherwise performance comparisons are not meaningful. Indeed, in the classical setting, one no longer compares optimal with non-optimal planners since both types of planners are very different: one provides guarantees that apply to all solutions, while the other provides guarantees that apply to one solution only. In the probabilistic setting this is even more subtle as there are different types of guarantees. For example, if we restrict ourselves to the class of MDPs that constitute the simplest generalization of the classical setting — the task of reaching the goal with certainty while minimizing the expected number of *steps* from a given initial state $s_0$ — there are methods that yield solutions (policies) that ensure that the goal will be reached with certainty in a finite number of steps (not necessarily optimal), and methods with no such guarantees. Both types of methods are necessary in practice, yet it is crucial to make a distinction among them and to identify useful benchmarks in each class. For methods that yield optimal policies, or at least policies with finite expected costs, standard dynamic programming methods like value iteration provide a useful bottom-line reference for assessing performance. In any case, we believe that useful benchmarks need to be defined taking into account the types of tasks that the various algorithms aim to solve, and the types of guarantees, if any, that they provide in their solutions.

GPT and mGPT are available for download at `http://www.ldc.usb.ve/~bonet`.

## Acknowledgements

mGPT was built upon a parser developed by John Asmuth from Rutgers University and Håkan Younes from Carnegie Mellon University. We also thank David E. Smith for comments that helped us to improve this note.